\documentclass[runningheads]{llncs}
\usepackage{graphicx, caption, tabularx}
\usepackage{multirow}
\usepackage{amsmath}
\usepackage{comment}
\usepackage{subcaption}
\usepackage{float}
\usepackage{xcolor} \usepackage{longtable}
\graphicspath{ {images/} }

\begin{document}
\title{Semantic WordRank: Generating Finer Single-Document Summarizations}
\author{Hao Zhang \and Jie Wang}
\institute{Department of Computer Science, University of Massachusetts, Lowell, MA, USA 
\email{hao\_zhang@student.uml.edu, jie\_wang@uml.edu}}
\maketitle

\begin{abstract}
We present Semantic WordRank (SWR), an unsupervised method for generating an extractive summary of a single document. Built on a weighted word graph with semantic and co-occurrence edges, SWR scores sentences using an article-structure-biased PageRank algorithm with a Softplus function adjustment, and promotes topic diversity using spectral subtopic clustering under the Word-Movers-Distance metric. We evaluate SWR on the DUC-02 and SummBank datasets and show that SWR produces better summaries than the state-of-the-art algorithms over DUC-02 under common ROUGE measures. We then show that, under the same measures over SummBank, SWR outperforms each of the three human annotators (aka. judges) and compares favorably with the combined performance of all judges.

\end{abstract}
 
\section{Introduction}

An extractive summarization algorithm of a single document aims to produce a much shorter text file to cover all the main points of the document without redundancy. Such an algorithm, unsupervised or supervised, extracts a few critical sentences from the document to form a summary.

An unsupervised method typically solves an optimization problem of selecting sentences subject to the underlying length constraint of a summary through sentence scoring and topic diversity. Unsupervised methods have three advantages: (1) They do not need training data; (2) they are in general independent of the underlying languages; and (3) they are much more efficient. Most unsupervised methods use easy-to-compute counting features, such as term frequency-inverse document frequency (TF-IDF) and co-occurrences of words, discarding semantic information. Some methods do incorporate semantic features including Semantic Role Labeling, WordNet, and Named-Entity Recognition when selecting sentences. These methods rely on specific language features which can be a potential problem when adapting across languages.

Supervised methods, on the other hand, may either require handcrafted features or learn feature representations automatically with a deep neural network model trained on a significant amount of data and is often time-consuming.

These considerations motivate us to investigate unsupervised summarization algorithms using semantic features that can be computed readily for any language. 

Word embedding representation maps a word into a high-dimensional vector representing continuous relations with other words, which is easy to obtain and does not rely on specific language features. A good pre-trained word embedding representation can help capture both useful semantic and syntactic information \cite{mikolov2013linguistic}. Word Mover's Distance (WMD) \cite{kusner2015word} utilizes this property to measure the distance between two documents, which brings semantics to sentence similarity. Our method, called Semantic WordRank (SWR), is an unsupervised graph-based algorithm with semantic features for generating extractive single-document summaries. It adds word-embedding representations as semantic relations to a graph with word co-occurrence relations. It then uses an article-structure-biased PageRank algorithm to compute a score for each word and a score for each sentence after an adjustment using the Softplus function. To facilitate coverage diversity, our approach uses a spectral subtopic clustering method to group subtopics under the WMD. Finally, it selects sentences with the highest scores over subtopic clusters in a round-robin fashion until the length constraint of the summary size fails.

The rest of the paper is organized as follows: We present in Section \ref{sec:2} a brief overview of related work on single-document extractive summarizations. Section \ref{sec:3} provide a detailed description of SWR. We describe in Section \ref{sec:4} our evaluation datasets, experiment settings, and comparison results under ROUGE measures on DUC-02 and SummBank datasets. We conclude the paper in Section \ref{sec:5}.

\section{Related Work} \label{sec:2}

\noindent \textbf{Unsupervised methods.} 
Relations between words play a central role in unsupervised extractive methods, with the underlying idea that related words ``promote'' each other. In particular, TextRank \cite{mihalcea2004textrank} and LexRank \cite{erkan2004lexrank} model a document as a sentence graph, and use the PageRank algorithm to rank sentences. TextRank offers robust performance over the DUC-02 dataset. These methods extract sentences based only on sentence scores, without considering topic diversity.

UniformLink \cite{wan2010exploiting} builds a sentence graph on a set of similar documents, where a sentence's score is computed based on both with-in document score and cross-document score. URank \cite{wan2010towards}  uses a unified graph-based framework to study both single-document and multi-document summarizations.

$E_{coh}$ \cite{parveen2015integrating}, as well as $T_{coh}$ \cite{parveen2015topical}, use a bipartite graph to represent a document and a different algorithm, Hyperlink-Induced Topic Search (HITS) \cite{kleinberg1998authoritative}, is used to score sentences. They both treat the summarization problem as an ILP problem, which maximizes the sentence importance, non-redundancy, and coherence at the same time. However, since ILP is an NP-hard problem, obtaining an exact solution to an ILP problem is intractable.

Submodularity optimization \cite{lin2011class} and Latent Semantic Analysis \cite{gong2001generic} are two other widely used unsupervised techniques for extractive summarizations.

\medskip

\noindent \textbf{Supervised methods.} 
Traditional supervised machine learning methods often need handcrafted features. These methods include Support Vector Machine and Naive Bayesian Classifier \cite{wong2008extractive}. CP3 \cite{parveen2016generating}, which has the same underlying idea as $E_{coh}$ and $T_{coh}$, mines coherence patterns in a corpus of abstracts, and extracts sentences by solving an ILP problem. 

Deep learning methods, able to learn sentence or document representations automatically, have recently been used to score sentences. For example, R2N2 \cite{cao2015ranking} uses a recursive neural network for both word level and sentence level scoring, followed by an ILP optimization strategy for selecting sentences. CNN-W2V  \cite{zhang2016extractive} is another example, which modifies a convolutional-neural-network (CNN) model of sentence classification \cite{kim2014convolutional} to rank sentences. SummaRuNNer \cite{nallapati2017summarunner}, on the other hand, treats summarizations as sequence classifications and uses a two-layer bi-directional recurrent neural network (RNN) model to extract sentences, where the first layer RNN is for words and the second layer is for sentences. Unlike unsupervised methods, the state-of-the-art deep learning approaches require a larger dataset and a significantly longer time to train a model, yet with a much lower ROUGE-1 score when evaluating on DUC dataset.

\section{Semantic WordRank} \label{sec:3}
We model a single-document extractive summarization problem as a 0-1 multi-objective knapsack problem.
Let $D$ be a document consisting of $n$ sentences indexed as $S_1, S_2, \ldots, S_n$
in the order they appear, each with a length $l_i$ and a score $s_i$, along with a maximum length capacity $L$, where $l_i$ is the number of characters contained in $S_i$. Let $F_{d}(D)$ denote a diversity coverage measure and $x_i$ a 0-1 variable
such that $x_i = 1$ if sentence $S_i$ is selected, and 0 otherwise. We model the summarization problem as the following multi-objective optimization problem:
\begin{equation}
\text{maximize} \sum_{i=1}^n s_ix_i \mbox{ and } F_{d}(D),~
\text{such that} \sum\limits_{i=1}^n l_ix_i \leq L \mbox{ and }x_i \in \{0, 1\}.
\end{equation}

In this section, we describe Semantic WordRank in details. In particular, we first describe how we score sentences, followed by a spectral subtopic clustering method to facilitate diversity. Finally, instead of solving the 0-1 knapsack problem to obtain an optimal solution, which is NP-hard, we use a greedy strategy to obtain an approximation.

\subsection{Semantic Word Graphs}

Let $G = (V, E)$ denote a weighted, undirected multiple-edge graph to represent a document $D$, where $V$ is a subset of words contained in $D$ that pass a part-of-speech filter, a stop-word filter, and a stemmer for reducing inflected words to the word stem. Two nodes in $G$ are connected if they co-occur within a window of $N$ successive words in the document, or the cosine similarity of their word-embedding representations exceeds a  threshold value $\Delta$. Let $u$ and $v$ be two adjacent nodes. Initially, they may have two types of connections, one for co-occurrence and one for semantic similarity. We assign the co-occurrence count of $u$ and $v$ as the initial weight to the co-occurrence connection, and the cosine similarity value as the initial weight to the semantic connection. Let $w_{u,v}^c$ and $w_{u,v}^s$ denote, respectively, the normalized weight for the co-occurrence connection and the semantic connection of $u$ and $v$. Finally, we assign $w_{u,v} = w_{u,v}^c+w_{u,v}^s$ as the weight to the incident edge of $u$ and $v$.

Note that the original TextRank algorithm \cite{mihalcea2004textrank} only considers co-occurrence between words. We add semantic similarity to represent that a pair of words express the same or similar meanings.

On a separate note, adding semantic similarity is vital for processing analytic languages, such as Chinese, which seldom use inflections. When stemming is not applicable, semantic similarity serves as an alternative to represent the relations between words with similar meanings, allowing them to share the importance when computing the PageRank scores for these nodes.

\subsection{Article-Structure-Biased PageRank}

Article structures define the order how to present information. For example, the typical structure of news articles is an inverted pyramid, presenting critical information at the beginning, followed by additional information with less important details. In academic writing, the structure would look like an hourglass, which includes an additional conclusion piece at the end of the article. To include article structures in our model, we use a position-biased PageRank algorithm \cite{florescu2017position}.

Let $G$ denote the word graph constructed in the previous section. The original TextRank algorithm computes score $W(v_i)$ of a node $v_i$ by iterating the following PageRank equation until converging, where $A(v_i)$ denotes the set of nodes adjacent to $v_i$:
	\begin{equation}
		W(v_i) =  \alpha \times \sum_{v_j \in A(v_i)} \frac{w_{ji}}{\sum\limits_{v_k \in A(v_j)} w_{jk}} W(v_j) + (1-\alpha),
	\end{equation}
	where $\alpha \in (0,1)$ is a damping factor, often set to 0.85 \cite{brin2012reprint}, and $w_{ji}$ is the weight of the edge between node $j$ and node $i$. The intuition behind this equation is that the importance of a node $v_i$ is related to the scores of its adjacent nodes and the probability $\beta = 1-\alpha$ of jumping from a random node to node $v_i$. In unbiased PageRank, each word is treated equally likely. In article-structure-biased PageRank, each word $v_i$ is biased with a probability $P(v_i)$ according to the underlying article structure. For example, in the inverted pyramid structure, we assign a higher probability to a word that appears closer to the beginning of the article.

We rank the importance of sentences from the most important to the least important based on the underlying article structure. In the inverted pyramid structure, we assign score $s_i$ to sentence $S_i$ using the reciprocal of indexing (or negative indexing). The probability for node $v_i$ can now be computed as follows:
	$$P(v_i) = \frac{C(v_i)}{\sum\limits_{v_j \in V} C(v_j)},~C(v_i) =\sum\limits_{v_i \in S_k} {s_k}.$$
	Note that the above computation is at the sentence level, which can be easily adapted to the word level by ranking words instead of sentences.

We compute the revised TextRank score $W'(v_i)$ for node $v_i$ as follows:
	\begin{equation}
		W'(v_i) =  \alpha \times \sum\limits_{v_j \in A(v_i)} \frac{w_{ji}}{\sum\limits_{v_k \in A(v_j)} w_{jk}} W'(v_j) + (1 - \alpha)\times P(v_i).
	\end{equation}
	We first assign an arbitrary value to each node and iterate the computation until it converges. The score associated with each node represents the word importance, and we refer it to as \textsl{salient score}.

\subsection{Sentence Scoring with Softplus Adjustment}

We use $W'(v_i)$ to compute a salient score for each sentence, where $v_i$ is a node in the word graph. At the first glance, one would sum up $W'(v_i)$ for words contained in sentence $S$ to be the salient score of $S$. This method has the following drawback. Suppose that two sentences $S_1$ and $S_2$ have similar salient scores with different $W'$ score distributions, where $S_1$ has a bipolar word score distribution where several keywords have very high scores and the rest have very low scores, while $S_2$ has roughly a uniform word score distribution. In this case, we would consider $S_1$ more critical than $S_2$, but using this method to compute a salient score of a sentence, we may end up with an opposite result.

To overcome this drawback, we use a Softplus Adjustment. The Softplus function $sp = \ln (1+e^x)$, commonly used as an activation function in neural networks, offers a significant enhancement when the input $x$ is a small positive number. When $x$ is large, we have $sp(x) \approx x$. We apply the Softplus Adjustment to each keyword, and then sum them up to get a salient score by
	\begin{equation}
		Salience_{sp}(S) = \sum\limits_{i=1}^k sp(salience_{w_i}).
	\end{equation}

\subsection{Spectral Clustering for Topic Diversity}
Spectral clustering \cite{von2007tutorial} uses eigenvalues of an affinity matrix to reduce dimensions before clustering. This model offers better performance over K-means with fewer requirements. The affinity matrix is an $n \times n$ square matrix, where each element corresponds to a similarity between two sentences. Since we use WMD metric, where smaller value between two sentences means more similar and larger value means less similar, we can use a RBF kernel to transfer WMD scores to similarity as follows: $sim(S_i, S_j) = e^{-\gamma * WMD(S_i, S_j)^2}$, where $\gamma$ can be set to 1.

Note that the time complexity for computing WMD is cubical in terms of the number of unique words in the sentences, making it difficult to scale. Linear-Complexity Relaxed WMD \cite{atasu2017linear} provides a faster approximation to WMD and we will use it to reduce time complexity.

We set the number of clusters $c_{num}$ based on the number of sentences $n$. According to our experiments, we find that 30\% of the original text size would be the best size for a summary to contain almost all significant points. To ensure 30\% summary size with no redundancy, we set $c_{num} = \min\{\lfloor0.3n\rfloor, 8\}.$

\subsection{Greedy Sentence Selections}
Each sentence $S_i$ now associates with four values: (1) sentence index $i$, (2) salient score $s_i$, (3) sentence length $l_i$, and (4) cluster index $c_j$ it belongs to. We select sentences 
in a round robin fashion as follows:
\vspace*{-6pt}
\begin{enumerate}
\item For each sentence $S_i$, compute the value per unit length using $s'_i = s_i/l_i$.
\item For each cluster, sort the sentences contained in it by their unit scores $s'_i$.
\item Group sentences with the highest unit scores, one for each cluster, to form a sentence sequence $S_c = \langle S_x, S_y, \ldots, S_{c_{num}}\rangle$. Select one sentence at a time from $S_c$ to the summary. Repeat this step while the size of the summary is less than $L$.
\end{enumerate}

\section{Experiments} \label{sec:4}
The DUC-02 \cite{duc2002} dataset contains a total of 567 news articles with an average of 25 sentences per document. Each article has at least two abstractive summaries written by human evaluators, where each summary is at most 100 words long. It has been a standard practice to use the DUC-02 benchmarks to evaluate the effect of summarization algorithms, including extractive summaries, even though DUC-02 benchmarks are abstractive summaries.

The SummBank \cite{radev2003summbank} dataset is a better dataset for evaluations, for it provides extractive summary benchmarks produced by human evaluators. SummBank offers, for each article, four extractive summaries, with a wide range of summary length from 5\% up to 90\% of, respectively, words and sentences. Three human judges annotate 200 news articles with an average of 20 sentences per document. In particular, given a document, each judge assigns an importance score to each sentence, resulting in, for a given length constraint, three extractive summaries, one for each judge, by selecting sentences with scores as high as possible and as relevant as possible. Moreover, summing up the three judges' scores for each sentence, we get a combined score and new summaries based on the new scores. We refer to the new summaries as a combined performance of all judges. 

ROUGE is the most used metric to evaluate the effect of summarizations. ROUGE-n is an n-gram recall between the automatic summary and a set of references, where ROUGE-SU4 evaluates an automatic summarization using skip-bigram and unigram co-occurrence statistics, allowing at most four intervening unigrams when forming skip-bigram.

\subsection{Setup}

\vspace*{-2pt}
We use a pre-trained word embedding representations with subword information \cite{bojanowski2016enriching} which can help solve the out-of-vocabulary problem and generate better word embedding for rare words. We run SWR (Semantic WordRank) to evaluate 30 articles selected uniformly at random from the DUC-01 dataset, and use the similarity value denoted by $\Delta$ that maximizes the ROUGE-1 score  as the threshold value for evaluations on DUC-02 and SumBank. We set $N=2$ as the window size for co-occurrences. Since both datasets contain only short news articles, it is reasonable to use inverted pyramid as the article structure.
\vspace*{-4pt}

\subsection{Evaluations}

\vspace*{-2pt}
On the DUC-02 dataset, we use SWR to extract sentences with a length of 100 words and use ROUGE to evaluate the results against the benchmarks. We can see from Table \ref{tab:1} that SWR outperforms the state-of-the-art algorithms under ROUGE-1 (R-1), ROUGE-2 (R-2), and ROUGE-SU4 (R-SU4), where UA means that the value is unavailable in the corresponding publications. 

\vspace*{-8pt}
\begin{table}[h]
\begin{center}
\begin{tabular}{l|c|c|c|c|c|c|c|c}
\hline 
\bf Methods \!&\! ULink \!&\! TextRank \!&\! $T_{Coh.}$ \!&\! URank \!&\! $E_{Coh.}$ \!&\! CNN-W2V  \!&\! CP3 \!&\! \bf SWR  \\
\hline
\bf R-1 & 47.1 & 47.1 & 48.1 & 48.5 & 48.5 & 48.6 & 49.0 & \bf 49.2 \\
\bf R-2 & 20.1 & 19.5 & 24.3 & 21.5 & 23.0 & 22.0 & \bf 24.7 & \bf 24.7 \\
\bf R-SU4 &UA & 21.7 & 24.2 & UA & 25.3 & UA & 25.8 & \bf 26.1 \\
\hline
\end{tabular}
\end{center}
\caption{Comparison results (\%) on DUC-02}
\label{tab:1}
\end{table}
\vspace*{-20pt}

We can see that SWR produces better R-1 and R-SU4 scores over those of CP3 and produces the same R-2 score as that of CP3. But more importantly, SWR offers higher adaptability across languages and lower time complexity. CP3 is a supervised method that requires annotated data to train coherence patterns, where entity recognition and co-reference resolution are needed. Moreover, CP3 generates summarization by solving a time-consuming ILP problem. On the contrary, SWR is an unsupervised model using only word embedding features that can be obtained easily for any language. 

\subsubsection{Comparisons with individual judges on SumBank}
\begin{figure}[H]
\centering
\begin{subfigure}{.5\textwidth}
  \centering
  \includegraphics[width=1\linewidth]{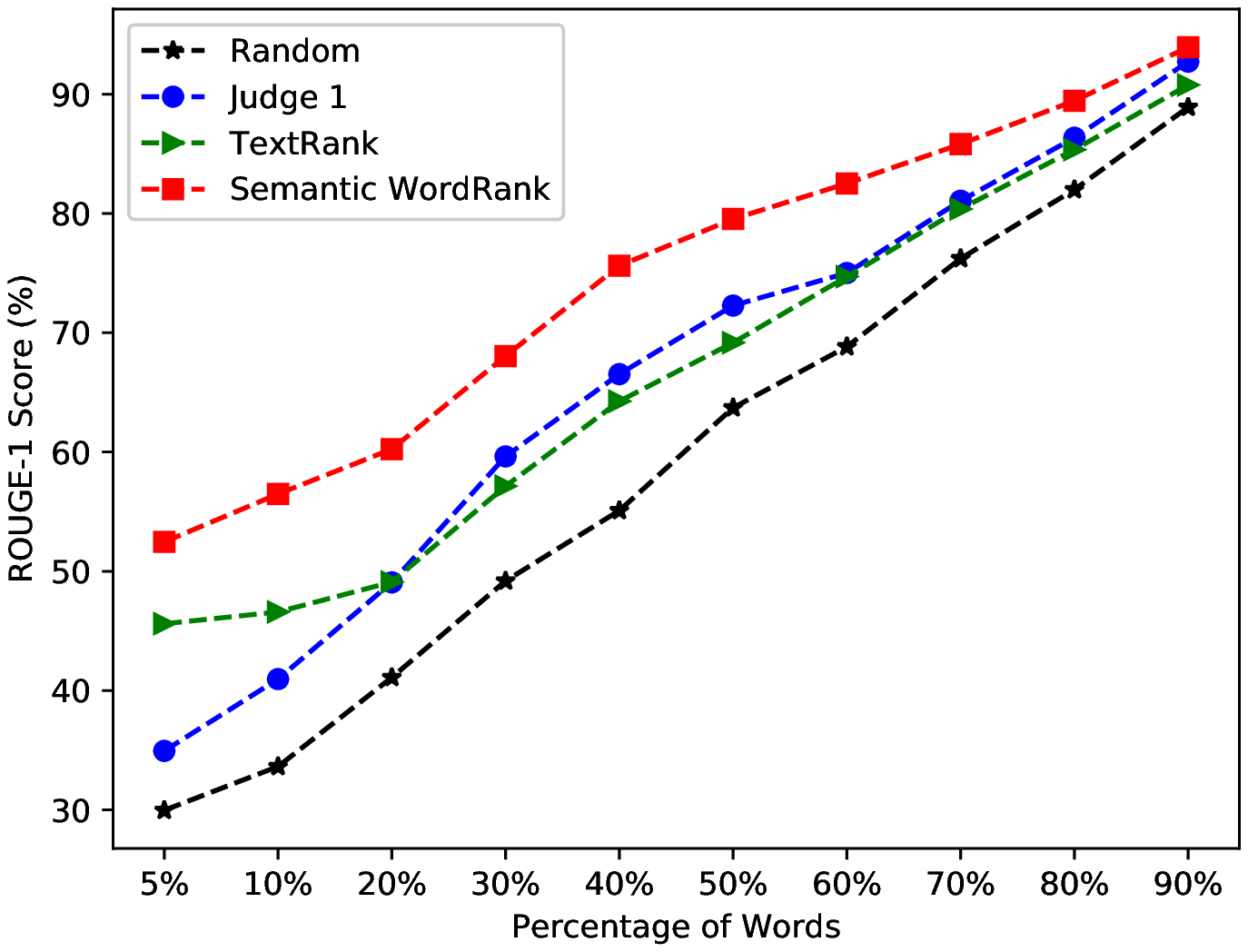}
   \caption{Judge 1 on words}
  \label{fig:f1_sub1}
\end{subfigure}%
\begin{subfigure}{.5\textwidth}
  \centering
  \includegraphics[width=1\linewidth]{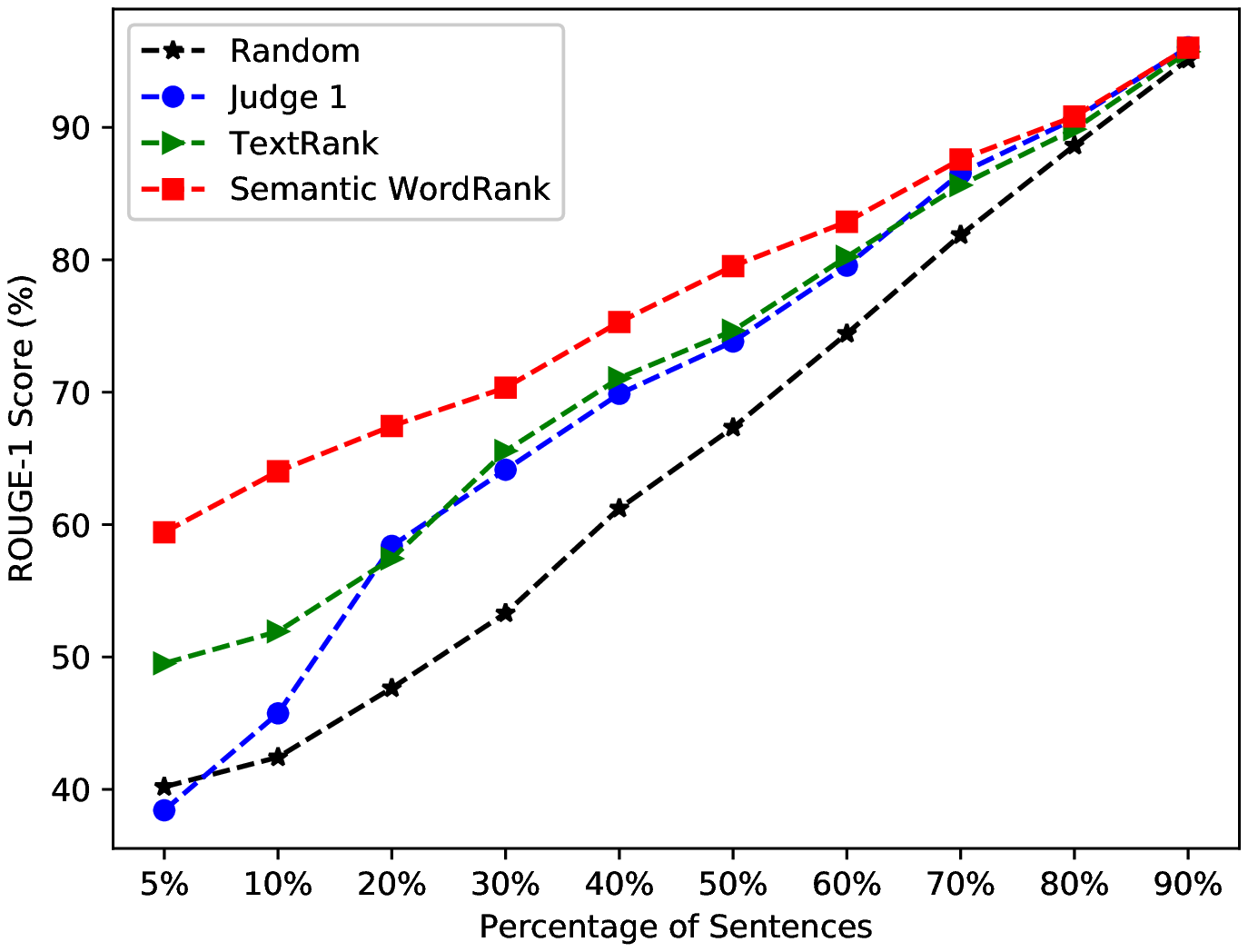}
  \caption{Judge 1 on sentences}
  \label{fig:f1_sub2}
\end{subfigure}
\begin{subfigure}{.5\textwidth}
\centering
\includegraphics[width=1\linewidth]{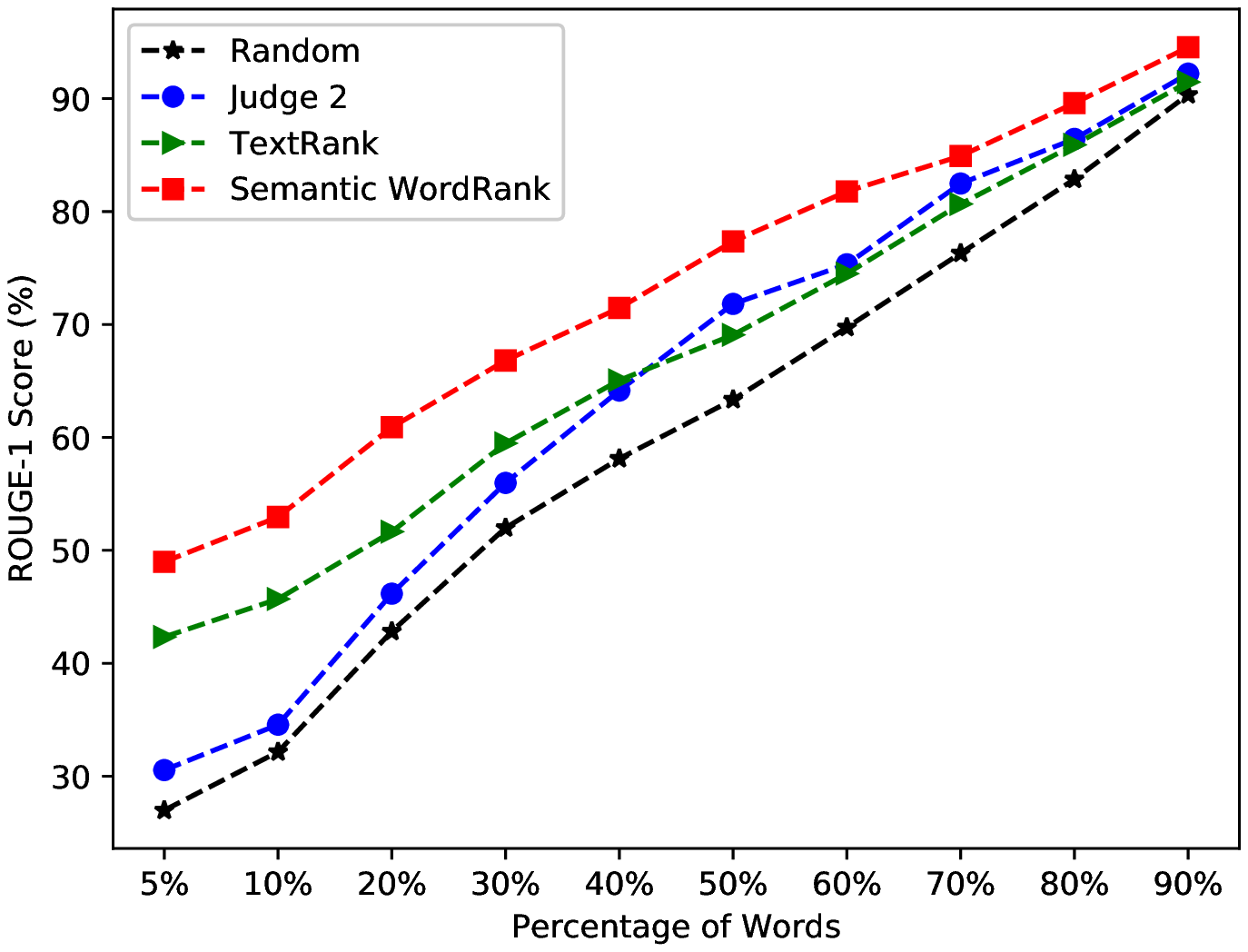}
\caption{Judge 2 on words}
\label{fig:f1_sub3}
\end{subfigure}%
\begin{subfigure}{.5\textwidth}
\centering
\includegraphics[width=1\linewidth]{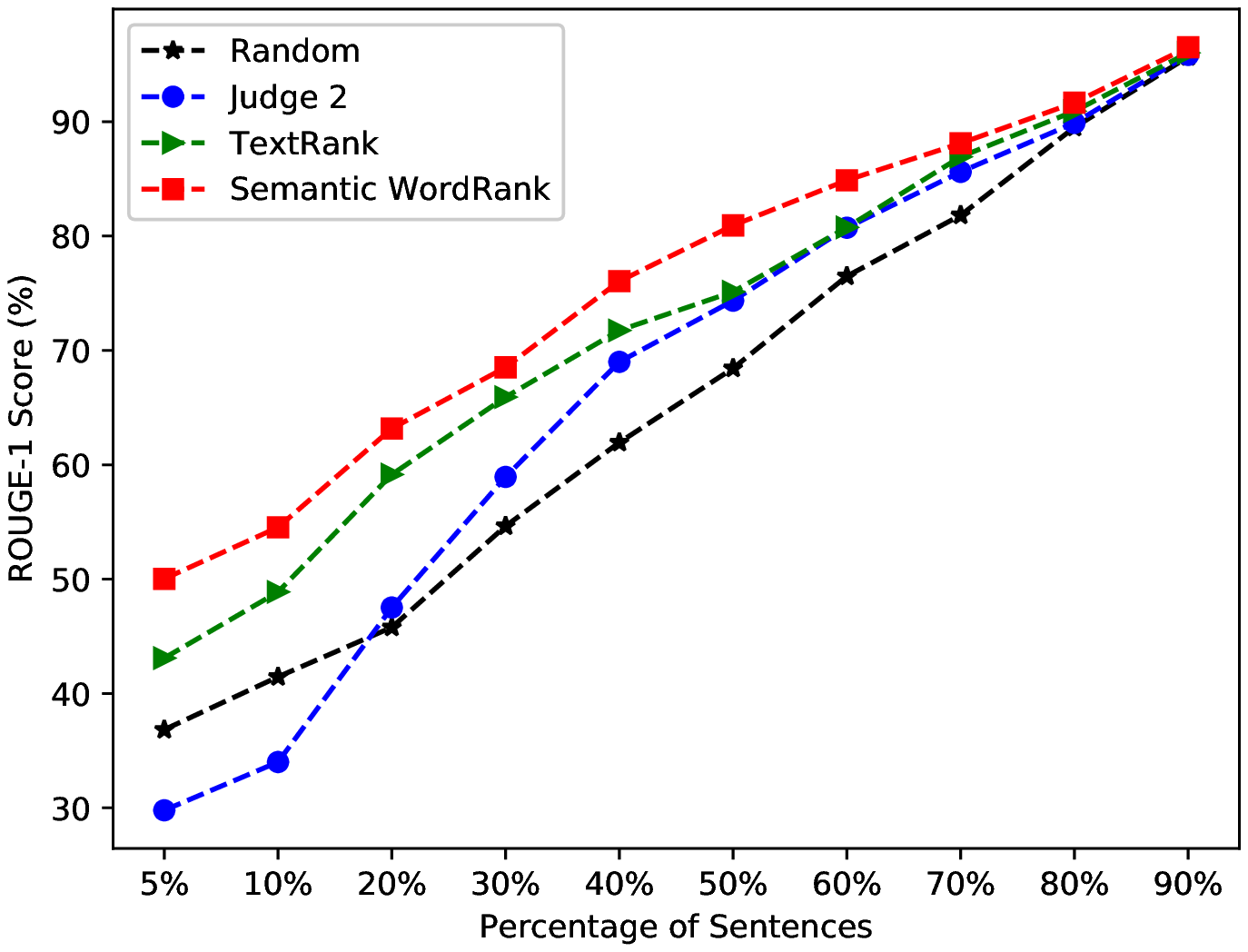}
\caption{Judge 2 on sentences}
 \label{fig:f1_sub4}
 \end{subfigure}
\begin{subfigure}{.5\textwidth}
 \centering
  \includegraphics[width=1\linewidth]{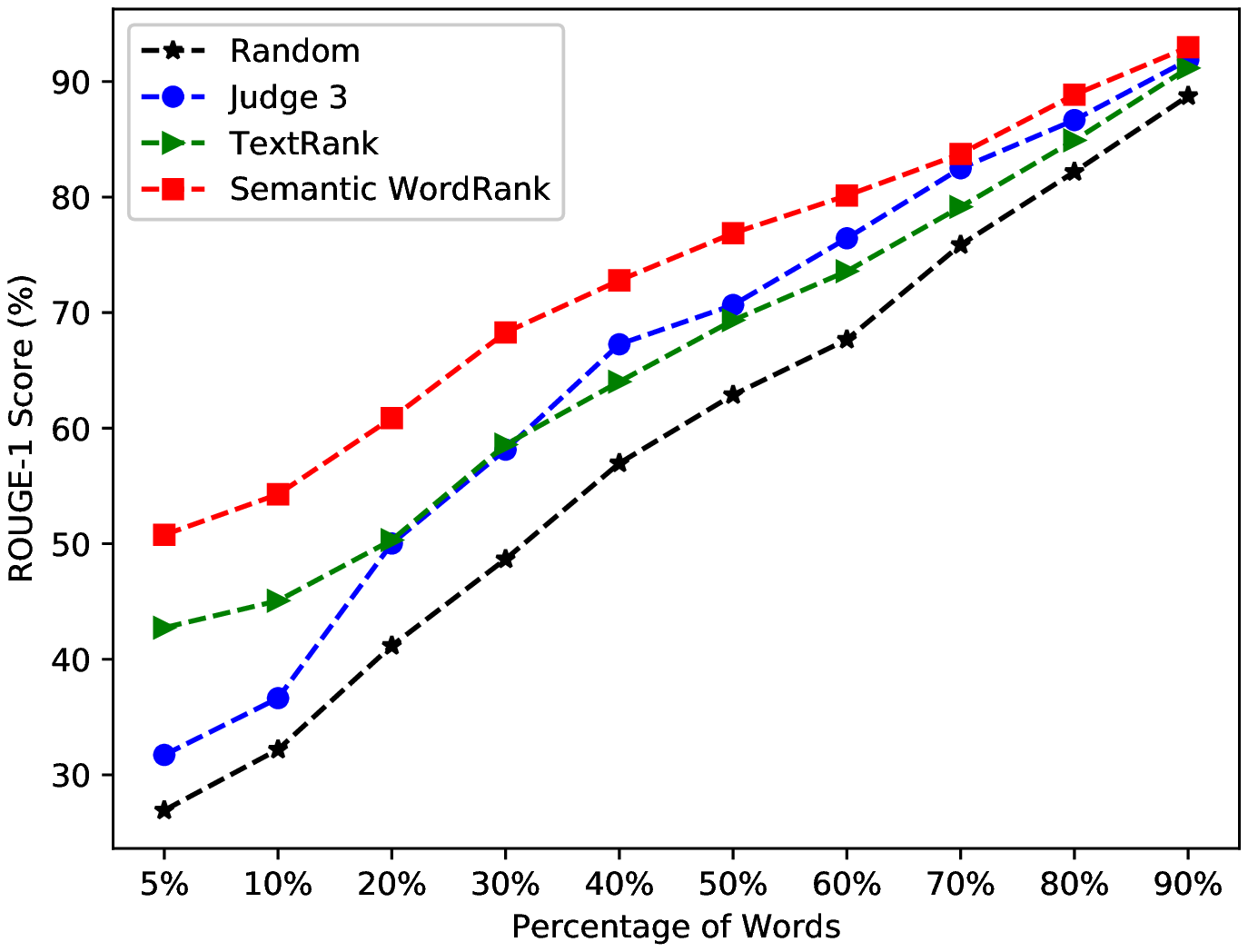}
    \caption{Judge 3 on words}
 \label{fig:f1_sub5}
\end{subfigure}%
\begin{subfigure}{.5\textwidth}
 \centering
 \includegraphics[width=1\linewidth]{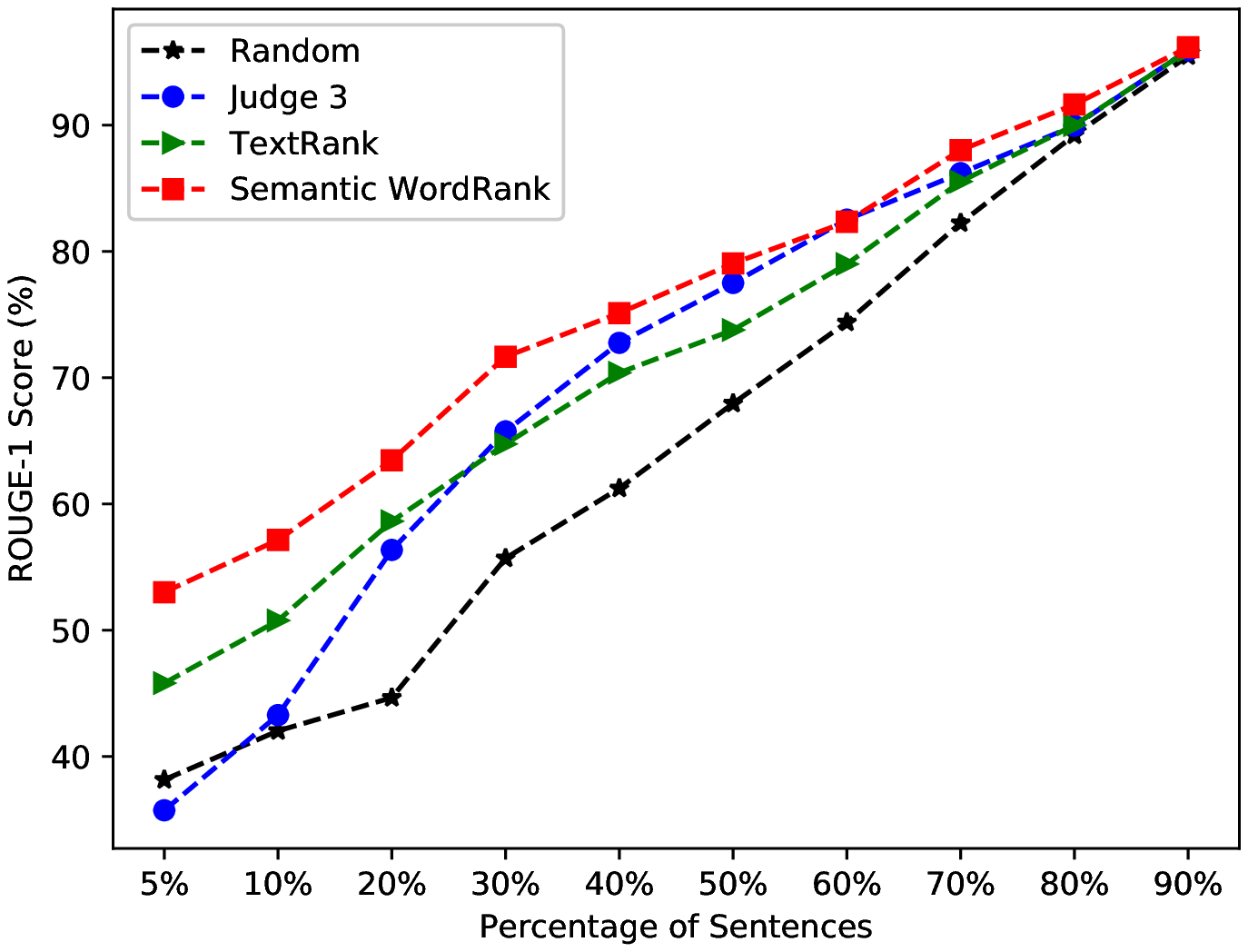}
 \caption{Judge 3 on sentences}
 \label{fig:f1_sub6}
\end{subfigure}
\caption{ROUGE-1 (\%) comparisons of SWR with individual judges, TextRank, and Random methods on the SummBank benchmarks}
\label{fig:1}
\end{figure}

To compare with each judge's summaries, for each judge $i$ ($i=1,2,3$), we use the other two judges' summaries as golden standards. Comparison with each judge are given in Figure \ref{fig:1}, where the words limitation or sentences limitation is from 5\% to 90\%. Table \ref{tab:2} depicts the ROUGE scores of different methods on summaries with 30\% of sentences and words, where S represents that the percentage is on the number of sentences and W on the number of words. We can see that while TextRank is comparable with individual judges, SWR significantly outperforms each judge.

\clearpage

\begin{center}
\begin{longtable}{l|c|c|c||c|c|c}
\hline \bf Methods \!&\! \bf R-1 (S) \!&\! \bf R-2 (S) \!&\! \bf R-SU4 (S) \!&\! \bf R-1 (W) \!&\! \bf R-2 (W) \!&\! \bf R-SU4 (W)\\ \hline
Random	& 55.70		&45.44		&43.69 &	49.19	&	37.53	&	36.44\\ \hline
\bf Judge 1	& 64.14		&57.11		&55.81	&	59.63	&	51.18	&	49.95\\
TextRank & 65.56		&58.28		&56.79	&	57.14	&	47.70	&	46.69\\
SWR	&\bf 70.35	&\bf64.47	& \bf 63.04	&	\bf68.15	&	\bf63.18	&	\bf61.92\\ \hline
\hline
Random	&	54.64	&	44.13	&	42.91 	&	51.99	&	40.67	&	39.42\\ \hline
\bf Judge 2	&	58.94	&	51.50	&	49.95	&	55.97	&	47.57	&	46.27 \\
TextRank	&	65.92	&	59.09	&	57.82 &	59.48	&	51.06	&	50.06\\
SWR	& \bf67.53	&\bf 61.45	&\bf 60.14 &	\bf66.11&	\bf59.08	&	\bf58.06\\ \hline
\hline
Random	&	55.70	&	45.44	&	43.69	&	48.67	&	37.07	&	36.08\\	\hline
\bf Judge 3	&	65.75	&	58.44	&	57.28	&	58.14	&	49.34	&	48.16\\
TextRank	&	64.75	&	57.32	&	55.70	&	58.58	&	49.72	&	48.53\\
SWR	&\bf 72.69	&\bf 67.65	&\bf 66.67	&	\bf69.32	&	\bf63.39	&	\bf62.14	\\ \hline
\caption{ROUGE (\%) comparisons of SWR individual judge, TextRank, and Random methods on SummBank with 30\% length constraint}
\label{tab:2}
\end{longtable}
\end{center}

\vspace*{-20pt}

\subsubsection{Comparisons with the combined performances of all judges}
\begin{figure}[H]
\centering
\begin{subfigure}{.5\textwidth}
  \centering
  \includegraphics[width=\linewidth]{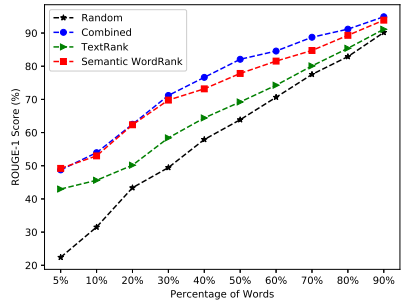}
  \caption{Different word constraints}
  \label{fig:f2_sub1}
\end{subfigure}%
\begin{subfigure}{.5\textwidth}
  \centering
  \includegraphics[width=\linewidth]{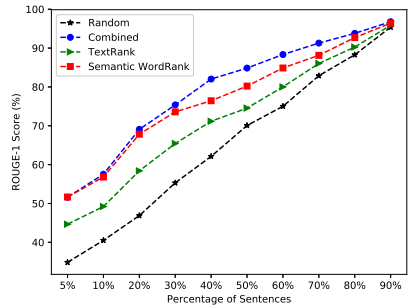}
  \caption{Different sentence constraints}
  \label{fig:f2_sub2}
\end{subfigure}
\caption{ROUGE-1 (\%) comparisons of SWR with the combined performance of all judges, TextRank, and Random methods on the SummBank benchmarks}
\label{fig:4}
\end{figure}

To compare SWR with the combined performance of all judges, we use the summaries of individual judges as golden standards. Figure \ref{fig:4} depicts the comparison results. We can see that SWR is compatible with the combined performances of all judges for summaries of length on and below 30\%, and close to the combined performance of all judges for more extended summaries.

\subsubsection{Other comparisons}

We investigate if semantic edge, Softplus function adjustment, article-structure-biased PageRank, and subtopic clustering work as expected. In particular, we remove each part one at a time to form a new method and evaluate it on the full Summbank dataset.

\begin{table}[h!]
\footnotesize
\begin{center}
\begin{tabular}{l|c|c|c|c|c|c|c|c|c}
\hline
\bf \multirow{2}{*}{Methods} & \bf R-1 & \bf R-2 & \bf R-SU4  & \bf R-1 & \bf R-2 & \bf R-SU4 & \bf R-1 & \bf R-2 & \bf R-SU4\\ \cline{2-10}
& \multicolumn{3}{c|}{10\%} & \multicolumn{3}{c|}{40\%} & \multicolumn{3}{c}{70\%}\\ \hline
SWR 	&	56.82	&	48.18	&	47.03   &76.48	&	71.74	&	70.39  &88.21	&	86.17	&	85.19\\
SWR\_NSE 	&	55.02	&	46.08	&	45.15  & \bf71.98	&	\bf66.76	&	\bf65.42	 &\bf85.94	&	\bf83.18	&	\bf81.83\\
SWR\_NAS 	&	\bf51.74	&	\bf40.87	&	\bf40.13  &73.52&	68.26	&	67.12  &87.82	&	84.95	&	84.12\\
SWR\_NSC 	&	56.36	&	47.34	&	46.44 & 75.68	&	70.35	&	69.03 & 87.54	&	84.49	&	83.54\\
SWR\_NSP	&	56.77	&	48.12	&	46.97 &	76.39	&	71.59	&	70.25 & 88.14	&	86.04	&	85.07\\
\hline
TextRank 	&	49.22	&	36.07	&	35.29 & 71.16	&	65.47	&	63.94 & 86.04	&	83.52	&	82.29\\
\hline
\end{tabular}
\end{center}
\caption{ROUGE (\%) comparison with different features removed}
\label{tab:3}
\end{table}

\vspace*{-20pt}

Table \ref{tab:3} depicts the evaluation results on 10\%, 40\%, and 70\% sentence constraints, where SWR\_NSE, SWR\_NAS, SWR\_NSC, and SWR\_NSP denote, respectively, SWR using no semantic edges, no article-structure information, no subtopic clustering, and no softplus adjustment. TextRank is added as a baseline model. We can see that when we remove a feature, the ROUGE scores drop, indicating that each part is essential. The numbers in bold are the biggest drops. We also see that when the summaries are short, removing article structure results in much larger drops, indicating that article structures are a more significant feature than the other two features. As summaries become longer, including semantic edges would become increasingly more critical. Our results indicate that using the Softplus function adjustment does improve the ROUGE scores, but not as significant as using the other three features.

\section{Conclusion and Future Work} \label{sec:5}
We present Semantic WordRank (SWR), an unsupervised method for generating an extractive summary of a single document. SWR incorporates semantic similarity to summarization while maintaining a high adaptability across languages. In particular, SWR generates an extractive summary using word embedding, Softplus function adjustment, article-structure-biased PageRank, and WMD spectral subtopic clustering. We evaluate SWR on DUC-02 and SummBank under three common ROUGE measures. Our experimental results show that, on DUC-02, our approach produces better results than the state-of-the-art methods under ROUGE-1, ROUGE-2, and ROUGE-SU4. We then show that, under the same ROUGE measures over SummBank, SWR outperforms each of the three individual human judges and compares favorably with the combined performance of all judges.

For a future project, we plan to adopt this approach to multi-document summarizations. We would also like to explore unsupervised sentence representations such as skip-thought vectors \cite{kiros2015skip} for summarization.

\section*{Acknowledgments}

We thank Liqun Shao for interesting conversations on using the Softplus function adjustment.

\bibliographystyle{splncs04}
\bibliography{Paper82_SWR}

\end{document}